\documentclass[letter]{ieice}
\usepackage{color}
\usepackage[fleqn]{amsmath}
\usepackage{newtxtext}
\usepackage[varg]{newtxmath}
\usepackage{multirow}
\usepackage{booktabs} 
\usepackage{cite}
\usepackage{mathtools}
\usepackage{subcaption}

\field{D}
\title{Kernel Logistic Regression Learning for High-Capacity Hopfield Networks}
\authorlist{%
  \authorentry{Akira Tamamori}{m}{AIT}\MembershipNumber{1512145}
  \affiliate[AIT]{The author is with the Department the
    Computer Science, Aichi Institute of Technology, Aichi, 470-0392, Japan.}
}
\received{2015}{1}{1}
\revised{2015}{1}{1}

\begin{document}
\maketitle

\begin{summary}
  Hebbian learning limits Hopfield network storage capacity
  (pattern-to-neuron ratio around 0.14). We propose Kernel Logistic
  Regression (KLR) learning. Unlike linear methods, KLR uses kernels
  to implicitly map patterns to high-dimensional feature space,
  enhancing separability. By learning dual variables, KLR dramatically
  improves storage capacity, achieving perfect recall even when
  pattern numbers exceed neuron numbers (up to ratio 1.5 shown), and
  enhances noise robustness. KLR demonstrably outperforms Hebbian and
  linear logistic regression approaches.
\end{summary}
\begin{keywords}
Hopfield network, Kernel logistic regression, Associative memory, Storage capacity, 
Noise robustness.
\end{keywords}

\section{Introduction}

Hopfield networks~\cite{Hopfield1982} provide a fundamental model for
content-addressable memory, capable of retrieving stored patterns from
noisy or incomplete cues through recurrent dynamics. These networks
operate by evolving their state over time to minimize an energy
function, with stored patterns corresponding to local minima
(attractors) of this function. However, the standard Hebbian learning
rule, which adjusts synaptic weights based on the correlation between
connected neurons' activities in the stored patterns, while simple and
biologically plausible due to its locality, severely limits the
network's storage capacity. . The theoretical limit is approximately
\( \beta = P / N \approx 0.14\) (where \(P\) is the number of
patterns and \(N\) is the number of
neurons)~\cite{Amit1985}. Exceeding this limit leads to the creation
of spurious attractors and interference between stored patterns,
resulting in catastrophic forgetting and retrieval errors.

Various methods have been proposed to improve capacity, often moving
beyond the constraints of simple Hebbian learning. One perspective is
to treat the learning process as finding weights that ensure each
stored pattern is a stable fixed point of the network dynamics. This
naturally leads to supervised learning formulations.  For each neuron
\(i\), the goal becomes predicting its correct state \(t_i^{\mu}\)
(representing the state \(\xi_i^{\mu}\) of neuron \(i\) in pattern
\(\mu\)) given the context of the pattern. Typically, the context is
the states of other neurons in \(\boldsymbol{\xi}^{\mu}\). Previous
work within this supervised framework has explored linear
approaches. For example, the pseudoinverse rule~\cite{Kohonen1989}
computes weights based on the Moore-Penrose pseudoinverse of the
pattern matrix, offering improved capacity over the Hebbian rule,
although it lacks locality. Alternatively, methods based on logistic
regression using linear predictors (Linear Logistic Regression - LLR)
have been considered~\cite{MacKay2002}, framing the task as a set of
independent linear classification problems for each neuron.

While these linear methods offer moderate improvements in capacity and
robustness, their effectiveness is still fundamentally limited by the
requirement of linear separability of the patterns, or more precisely,
the linear separability required for each neuron's prediction task. To
capture more complex relationships between patterns and potentially
achieve substantially greater capacity, we turn to the power of
non-linear kernel methods. Kernel Logistic Regression (KLR) is a
technique that allows us to implicitly map the input patterns into a
high-dimensional (potentially infinite-dimensional) feature space
using a kernel function~\(K(\cdot, \cdot)\). The ``kernel
trick''~\cite{Scholkopf2001} enables us to compute dot products and
perform linear operations like logistic regression in this feature
space without explicitly computing the high-dimensional mapping
\(\Phi(\cdot)\), making computations feasible. The intuition is that
patterns that are not linearly separable in the original input space
might become linearly separable in this richer feature space. KLR
learns dual variables associated with each stored pattern, effectively
finding a non-linear decision boundary for each neuron's prediction
task.

In this letter, we implement and rigorously evaluate a KLR-based
learning algorithm specifically tailored for Hopfield networks. We
employ the Radial Basis Function (RBF) kernel. Through systematic
simulations, we compare the performance of KLR learning against both
the traditional Hebbian rule and the Linear Logistic Regression (LLR)
approach. Our results demonstrate that significant gains achieved by
KLR in both storage capacity and noise robustness, establishing it as
a potent method for enhancing the capabilities of Hopfield-like
associative memory systems.

\section{Methods}
\subsection{Model Setup}
We consider a network of \(N\) bipolar neurons \(\{-1, 1\}^{N}\). Let
\(\{\boldsymbol{\xi}^{\mu}\}_{\mu=1}^{P}\) be the set of \(P\)
patterns to be stored. For KLR training, we transform these into
target vectors \( \mathbf{t}^{\mu} \in \{0, 1\}^{N}\) where
\(t_i^{\mu} = (\xi_i^{\mu} + 1) / 2\).  The network state
\(\mathbf{s}(t) \in \{-1, 1\}^{N}\) evolves over discrete time steps.

\subsection{Learning Algorithms}

\subsubsection{Hebbian Learning (Baseline)}
The standard Hebbian weight matrix \(\mathbf{W}^{\text{Heb}}\) is
computed as
\(\mathbf{W}^{\text{Heb}} = (1/N) \mathbf{X}^{\top} \mathbf{X} \) with
diagonal elements set to zero, where
\(\mathbf{X} = \left[ \boldsymbol{\xi}^{1}, \ldots
  \boldsymbol{\xi}^{P}\right]^{\top}\).

\subsubsection{Linear Logistic Regression (LLR)}
For comparison, we consider LLR where each neuron \(i\)
learns a weight vector \(\mathbf{w}_i^{\text{LLR}}\) (which forms part of an \(N \times N\) weight matrix \(\mathbf{W}^{\text{LLR}}\)) to
predict \(t_i^{\mu}\) from \(\boldsymbol{\xi}^{\mu}\) (excluding self-connection).
The logit is given as:
$$
h_i^{\nu} = \sum_{j = 1, j\neq i}^{N} \mathbf{w}_{ij}^{\text{LLR}} \xi_j^{\nu}
$$
where \(\mathbf{w}_{ij}^{\text{LLR}}\) is the synaptic weight from neuron \(j\) to neuron \(i\), and \(\xi_j^{\mu}\) is the state of neuron \(j\) in pattern \(\mu\). 
Weights are learned by minimizing the regularized negative log-likelihood via gradient descent~\cite{MacKay2002}. The resulting \(N \times N\) weight matrix \(\mathbf{W}^{\text{LLR}}\) (symmetrized) is used for recall.

\subsubsection{Kernel Logistic Regression Learning}
The core idea is to train each neuron $i$ independently using KLR, learning dual variables $\boldsymbol{\alpha}_i = [\alpha_{1i}, \dots, \alpha_{Pi}]^T$ instead of primal weights. Let $\boldsymbol{\alpha}$ be the \(P \times N\) matrix whose $i$-th column is $\boldsymbol{\alpha}_i$.
The logit for neuron \(i\) given pattern \(\boldsymbol{\xi}^{\nu}\) is:
$$
h_i^{\nu} = \sum_{\mu=1}^P K(\boldsymbol{\xi}^{\nu}, \boldsymbol{\xi}^{\mu}) \alpha_{\mu i}
$$
We use the RBF kernel \(K(\mathbf{x}, \mathbf{y}) = \exp(\gamma \| \mathbf{x} - \mathbf{y}
\|^{2})\). The predicted probability is given by the logistic sigmoid function:
$$
y_i^{\nu} = \sigma(h_i^{\nu}) = \frac{1}{1 + \exp(-h_i^{\nu})}
$$
The learning objective is minimizing the negative log-likelihood with
L2 regularization on \(\boldsymbol{\alpha}_i\):
$$
L(\boldsymbol{\alpha}_i) = -\sum_{\nu=1}^P [ t_i^{\nu} \log(y_i^{\nu}) + (1 - t_i^{\nu}) \log(1 - y_i^{\nu}) ] + \frac{\lambda}{2} \boldsymbol{\alpha}_i^{\top} \mathbf{K} \boldsymbol{\alpha}_i, 
$$
where \(\mathbf{K}\) is the \(P \times P \) Gram matrix (\(\mathbf{K}_{\nu \mu} = K(\boldsymbol{\xi}^{\nu}, \boldsymbol{\xi}^{\mu}) \)). Minimizing \(L\) with respect to \( \boldsymbol{\alpha}_i \) using gradient descent involves the gradient:

$$
\frac{\partial L}{ \partial \boldsymbol{\alpha}_i} = \mathbf{K} (\mathbf{y}_i - \mathbf{t}_i + \lambda \boldsymbol{\alpha}_i)
$$
where \(\mathbf{y}_i = [y_{i}^{1}, \ldots, y_{i}^{P}]^{\top}\) and \(\mathbf{t}_i = [t_{i}^{1}, \ldots, t_{i}^{P}]^{\top}\).

\subsection{Recall Process}
The state update rule differs between models:

\begin{description}
\item[Hebbian and LLR:] \(s_{i}(t) = \text{sign}(\sum_{j\neq i} \mathbf{W}_{ij} s_{j}(t)) \)  (using \(\mathbf{W}^{\text{Heb}}\) or \(\mathbf{W}^{\text{LLR}}\)), where
  \(s_{i}(t)\) is the \(i\)-th element of the state \(\mathbf{s}(t)\).
\item [KLR:]  Update without an explicit \(N \times N\) weight matrix \(\mathbf{W}\):
\begin{enumerate}
\item Compute kernel values: \\\(\mathbf{k}_{\mathbf{s}(t)} = \left[ K(\mathbf{s}(t), \boldsymbol{\xi}^{1}) , \ldots, K(\mathbf{s}(t), \boldsymbol{\xi}^{P})\right]\) (size \(1 \times P\)).
\item Compute the logit vector for all neurons:\\
$\mathbf{h}(\mathbf{s}(t)) = [h_1(\mathbf{s}(t)), \dots, h_N(\mathbf{s}(t))]$ (size $1 \times N$),  where the $i$-th component $h_i(\mathbf{s}(t))$ is computed as:
      \[ h_i(\mathbf{s}(t)) = \sum_{\mu=1}^{P} K(\mathbf{s}(t), \boldsymbol{\xi}^\mu)\alpha_{\mu i}. \]
      This can be expressed in matrix form as $\mathbf{h}(\mathbf{s}(t)) = \mathbf{k}_s(t)\boldsymbol{\alpha}$.
\item Update the state: \(\mathbf{s}(t+1) = \text{sign}(\mathbf{h}(\mathbf{s}(t)) - \boldsymbol{\theta})\), where \(\boldsymbol{\theta}\) is a threshold vector.
\end{enumerate}
\end{description}
It is noted that KLR recall involves kernel computations with all \(P\) stored patterns at each step.

\section{Experiments}
\subsection{Experimental Setup}
We simulated networks with \(N = 500\) neurons.  Random bipolar
patterns were generated  with \(P(\xi_{i}^{\mu} = 1) = 0.5\).
We compared Hebbian, LLR, and KLR (RBF kernel, \(\gamma =1/N\)).
For LLR and KLR, learning parameters were: regularization
\(\lambda=0.01\), learning rate \(\eta=0.1\) (the step
  size for the gradient descent optimization), and number of updates
= 200.  Recall dynamics were tracked for \(T=25\) steps.  Overlap
\(m(t)=(1/N) \mathbf{s}(t)^{\top}\boldsymbol{\xi}^{\text{target}} \)
was measured.  Recall was successful if \(m(T) > 0.95\).

\subsection{Storage Capacity Evaluation}
Recall was considered successful if the final overlap between the
network state and the target pattern exceeded 0.95.  We measured the
success rate starting from the original patterns
(\(\mathbf{s}(0) = \boldsymbol{\xi}^{\mu}\)) as a function of storage
load \(\beta = P/N\), starting from the original patterns.  Figure 1
presents the results for networks with \(N=500\) neurons.  The Hebbian
network's performance collapses around \(\beta \approx 0.14\),
consistent with theoretical predictions~\cite{Amit1985}.  LLR offers
substantial improvement, maintaining high success rates up to
\(\beta \approx 0.85\) before declining sharply and failing completely
by \(\beta = 0.95\).  Strikingly, Kernel Logistic Regression (KLR)
dramatically outperforms both, achieving and maintaining a 100\%
success rate throughout the entire tested range, up to
\(\beta = 1.5\). This demonstrates its ability to stably store and
recall patterns even when the number of patterns significantly exceeds
the number of neurons (\(P > N\)).
\begin{figure}[t]
\begin{center}
  \includegraphics[width=0.97\hsize]{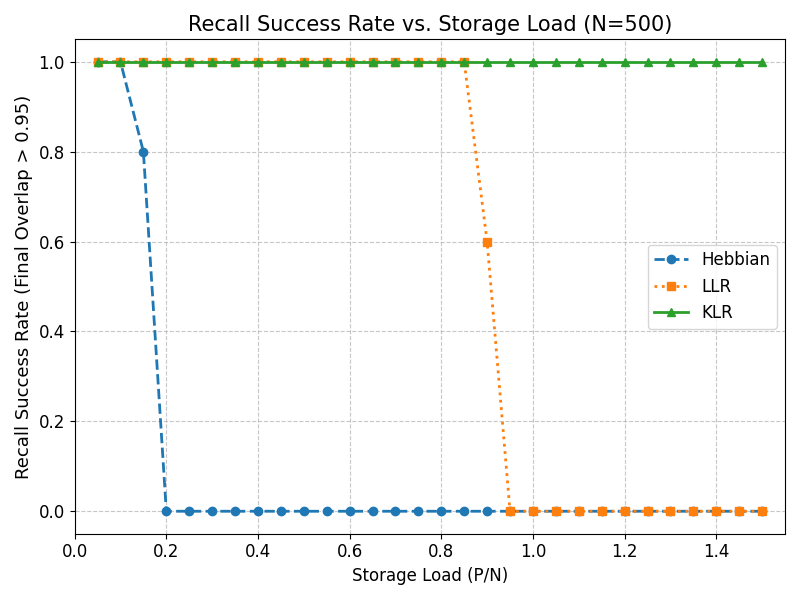}
\caption{Recall Success Rate vs. Storage Load for Hebbian, LLR, and KLR.}
\end{center}
\end{figure}

\subsection{Noise Robustness Evaluation}
We evaluated the final overlap with the target pattern as a function
of the initial overlap, \(m(0)\), for a fixed intermediate load
(\(\alpha = 0.2\), \(N=500\), \(P=100\)).  Initial states
\(\mathbf{s}(0)\) were generated by flipping a fraction \((1-m(0))/2\)
of bits in the target pattern.
Figure 2 plots the mean final overlap \(m(T)\) achieved after \(T=25\)
steps against \(m(0)\).  The Hebbian network consistently failed to
recall the pattern accurately, with the final overlap remaining low
(approximately $0.2$--$0.35$) even for high initial overlaps (e.g.,
\(m(0)=0.9\)). LLR demonstrated improved robustness, achieving
successful recall (\(m(T) \approx 1.0\)) when the initial overlap was
greater than approximately \(m(0) = 0.4\). KLR exhibited significantly
superior robustness, reaching perfect recall (\(m(T) = 1.0\)) from
initial states with much lower overlap, starting around
\(m(0) = 0.2\). This indicates that KLR possesses a considerably
larger basin of attraction compared to both Hebbian learning and LLR.
\begin{figure}[t]
\begin{center}
  \includegraphics[width=0.97\hsize]{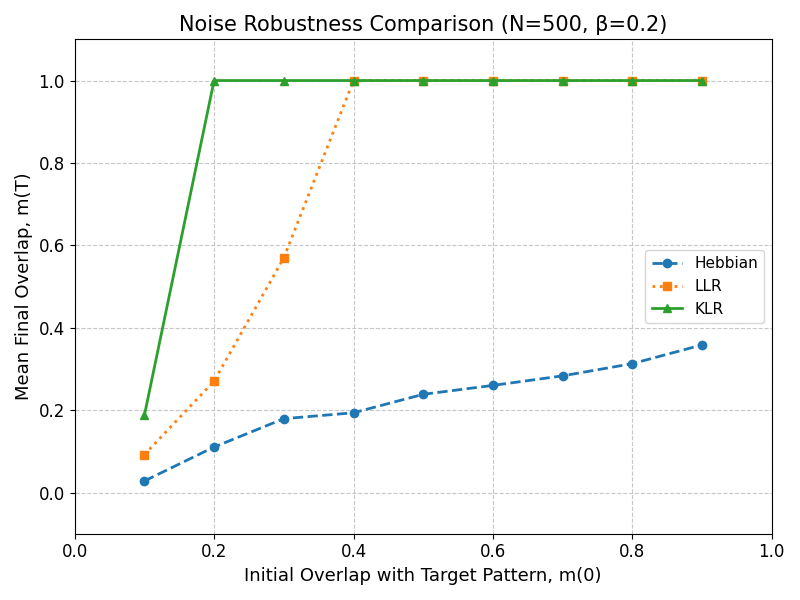}
\caption{Final Overlap $m(T)$ vs. Initial Overlap $m(0)$ at $\beta=0.2$ for Hebbian, LLR, and KLR.}
\end{center}
\end{figure}

\subsection{Effect of Kernel Parameter \(\gamma\)}
To investigate the influence of the RBF kernel width, we evaluated KLR
performance at a fixed load \(\beta = 0.3\) (\(N=500\), \(P=150\)) for
different \(\gamma\) values, scaled relative to \(1/N\). Figure 3
shows the recall success rate against the scaled parameter
$\tilde{\gamma} = \gamma N$.  The results demonstrate that performance
is sensitive to the choice of \(\gamma\). When \(\gamma\) was too
small (\(\tilde{\gamma}\) = 0.1 and 0.5), the network failed to recall the
patterns, yielding a success rate of 0.0. This suggests that an overly
broad kernel fails to effectively separate patterns in the feature
space. However, for \(\tilde{\gamma} \ge 1.0\) (corresponding to
\(\gamma \ge 1/N\)), the network consistently achieved a 100\% success
rate within the tested range (up to \(\tilde{\gamma} = 10.0\)). In this
experimental setting, performance did not degrade even with larger
\(\gamma\) values. This indicates that \(\gamma = 1/N\), our chosen
default value for other experiments, is a reasonable and effective
choice, falling within the range of successful parameter values.
\begin{figure}[t]
\begin{center}
  \includegraphics[width=0.97\hsize]{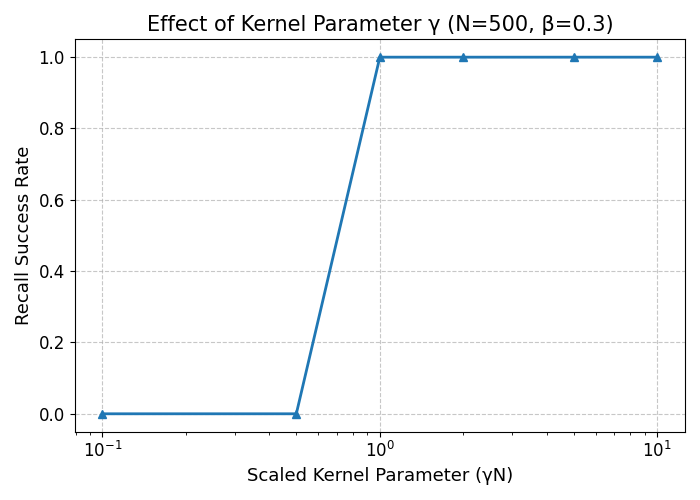}
\caption{Recall Success Rate vs. scaled factor ($\gamma N$) at $\beta=0.3$ for KLR.}
\end{center}
\end{figure}

\subsection{Effect of Regularization Parameter \(\lambda\)}
\begin{figure}[t]
\begin{center}
  \includegraphics[width=0.97\hsize]{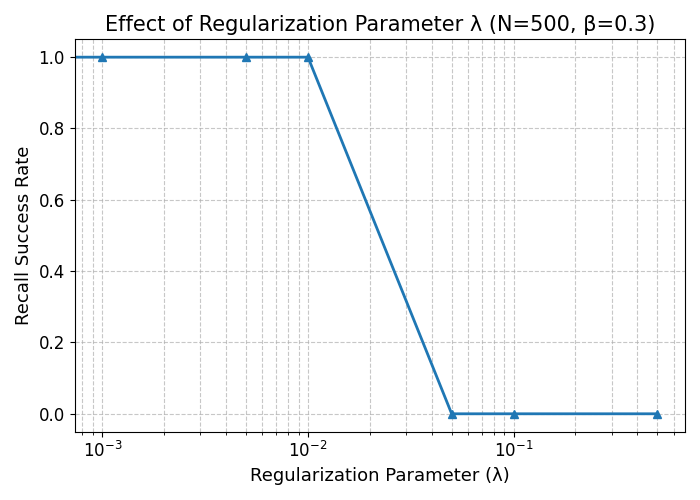}
\caption{Recall Success Rate vs. $\lambda$ at $\beta=0.3$ for KLR.}
\end{center}
\end{figure}
We also examined the effect of the L2 regularization parameter
\(\lambda\) on KLR performance at at a fixed load \(\beta=0.3\)
(\(N=500\), \(P=150\)) with \(\gamma=1/N\).  Figure 4 plots the
success rate against \(\lambda\).  The results show that the network
achieved a 100\% success rate for \(\lambda\) values ranging from 0
(no regularization) up to 0.01. This suggests that mild regularization
in this range does not impair recall performance under these
conditions, and potentially offers some benefit in terms of learning
stability or generalization (though not directly tested
here). However, when \(\lambda\) was increased further to 0.05 or
higher, the success rate dropped abruptly to 0.0. This indicates that
excessive regularization hinders the learning process significantly,
preventing the network from adequately fitting the patterns, likely
due to overly constrained dual variables. Our chosen default value,
\(\lambda=0.01\), falls within the effective range and avoids the
performance degradation caused by stronger regularization.

\section{Discussion}
Our experiments with $N=500$ neurons clearly demonstrate that Kernel
Logistic Regression (KLR) dramatically enhances Hopfield network
performance, not only in terms of noise robustness but especially in
storage capacity. The capacity improvement is remarkable; KLR achieved
perfect recall throughout the entire tested range, up to a storage
load of \(\beta=1.5\) (Fig. 1). This vastly exceeds the classic
Hebbian limit (\(\beta \approx 0.14\)) [2] and significantly surpasses
LLR, which failed around \(\beta \approx 0.9\). The ability of KLR to
successfully store and recall patterns even when the number of
patterns significantly exceeds the number of neurons (\(P > N\)) is
particularly noteworthy and highlights its powerful capabilities. This
enhanced robustness is also significant, with KLR networks forming
considerably larger basins of attraction (Fig. 2), successfully
recalling patterns from initial states with much lower overlap
(\(m(0) \approx 0.2\)) compared to LLR (\(m(0) \approx 0.4\)) and the
Hebbian rule.

This superior performance is likely attributable to KLR's ability to
leverage the high-dimensional (potentially infinite-dimensional)
feature space implicitly defined by the RBF kernel. This allows it to
learn complex, non-linear decision boundaries for each neuron,
enabling effective pattern separation even when patterns are densely
packed or linearly inseparable in the original input space. The fact
that KLR operates flawlessly well into the $P > N$ regime, where input
patterns are necessarily linearly dependent, strongly suggests that
patterns remain effectively separable within the kernel-induced
feature space. This contrasts sharply with linear methods like LLR,
whose performance degrades when linear separability becomes
challenging. The superior noise robustness further underscores the
effectiveness of these non-linear boundaries.

As expected for kernel methods, performance depends on hyperparameter
choices. Our investigation into the RBF kernel parameter \(\gamma\)
(Fig. 3) revealed that performance is poor when the scaled parameter
$\tilde{\gamma} = \gamma N$ is less than 1.0, but optimal performance
was achieved and maintained for $\tilde{\gamma} \ge 1.0$ within the
tested range (up to $\tilde{\gamma}=10.0$).  This confirms that
\(\gamma=1/N\) ($\tilde{\gamma} = 1.0$), our chosen default, is a
reasonable and effective choice. Similarly, the L2 regularization
parameter \(\lambda\) (Fig. 4) showed optimal performance for
\(\lambda\) between 0 and 0.01, with a sharp drop for
\(\lambda \ge 0.05\), confirming that mild or no regularization is
appropriate here.

This work aligns with and complements the theoretical framework of
``Kernel Memory Networks''~\cite{Iatropoulos2022}. While
\cite{Iatropoulos2022} primarily focused on kernel SVM and optimal
robustness bounds, our work provides concrete empirical validation for
the KLR formulation, demonstrating its practical effectiveness and its
potential to achieve storage capacities well beyond the number of
neurons.

However, a crucial practical consideration is the computational
cost. Kernel Logistic Regression learning, particularly when the
number of patterns \(P\) is large, can be computationally intensive. This
cost stems primarily from handling the \(P \times P\) kernel matrix
during learning (involving \(O(N^{2})\) computation or memory) and the
recall process, which requires $P$ kernel evaluations followed by
matrix operations (roughly \(O(PN)\) complexity per step, see Sec
2.3). This contrasts with the \(O(N^{2})\) recall complexity of
Hebbian or LLR (assuming precomputed weights) and presents a
scalability challenge, particularly as $P$ approaches $N$. Our own
simulation experiences confirmed that computation time increases
significantly with $N$ and $P$.  This trade-off between the
demonstrated high performance and computational demands underscores
the necessity of exploring efficient kernel approximation methods,
such as the Nystr{\"o}em technique~\cite{Williams2000}, to make KLR
feasible for larger-scale network applications.

Future directions include validating the effectiveness of
approximations like Nyströem, evaluating other kernel types (e.g.,
polynomial), developing efficient online KLR learning rules, and
perhaps a more detailed analysis comparing the effects of different
regularization forms (\(\|\boldsymbol{\alpha}\|^{2} \) and
\(\boldsymbol{\alpha}^{\top} \mathbf{K}\boldsymbol{\alpha}\)).

\section{Conclusion}
We have demonstrated that Kernel Logistic Regression provides a
powerful learning mechanism for Hopfield networks, substantially
increasing storage capacity and noise robustness compared to
traditional Hebbian learning and linear logistic regression. By
leveraging kernel methods to implicitly perform non-linear feature
mapping, KLR enables more effective pattern separation and
recall. Despite the increased computational cost associated with
kernel evaluations, the significant performance gains make KLR a
compelling approach for building high-performance associative memory
systems. This work highlights the potential of applying modern
kernel-based machine learning techniques to enhance classic neural
network models.

\bibliographystyle{ieicetr}
\bibliography{paper}

\begin{thebibliography}{1}

\bibitem{Hopfield1982}
J.J. Hopfield, ``Neural networks and physical systems with emergent collective
  computational abilities.,'' Proceedings of the National Academy of Sciences,
  vol.79, no.8, pp.2554--2558, 1982.

\bibitem{Amit1985}
D.J. Amit, H.~Gutfreund, and H.~Sompolinsky, ``Storing infinite numbers of
  patterns in a spin-glass model of neural networks,'' Phys. Rev. Lett.,
  vol.55, pp.1530--1533, Sep\ 1985.

\bibitem{Kohonen1989}
T.~Kohonen, Self-organization and associative memory: 3rd edition,
  Springer-Verlag, Berlin, Heidelberg, 1989.

\bibitem{MacKay2002}
D.J.C. MacKay, Information Theory, Inference \& Learning Algorithms, Cambridge
  University Press, USA, 2002.

\bibitem{Scholkopf2001}
B.~Scholkopf and A.J. Smola, Learning with Kernels: Support Vector Machines,
  Regularization, Optimization, and Beyond, MIT Press, Cambridge, MA, USA,
  2001.

\bibitem{Iatropoulos2022}
G.~Iatropoulos, J.~Brea, and W.~Gerstner, ``Kernel memory networks: a unifying
  framework for memory modeling,'' Advances in Neural Information Processing
  Systems, 2022.

\bibitem{Williams2000}
C.K.I. Williams and M.~Seeger, ``Using the nystr\"{o}m method to speed up
  kernel machines,'' Advances in Neural Information Processing Systems, 2000.

\end{thebibliography}
\end{document}